\documentclass[10pt,twocolumn,letterpaper]{article}

\usepackage{cvpr}              

\usepackage{CJK}
\usepackage{multicol}
\usepackage{color}
\usepackage{xcolor}
\usepackage{colortbl}
\usepackage{bm}
\usepackage{extarrows}
\usepackage{float}
\usepackage{bbm}
\usepackage{booktabs}
\usepackage{multirow}
\usepackage{amsmath}
\usepackage{amssymb}
\usepackage{algorithm}
\usepackage{algorithmic}
\usepackage{makecell}
\usepackage{amsmath}
\usepackage{amssymb}
\usepackage{amsthm}
\usepackage[utf8]{inputenc}
\usepackage{amsfonts} 
\usepackage{caption} 
\usepackage{newfloat}
\usepackage{listings}
\usepackage{epsfig}
\usepackage{graphicx}
\usepackage{amsmath}
\usepackage{amssymb}
\usepackage{booktabs}
\usepackage{multirow}
\usepackage{makecell}
\usepackage{xcolor}
\usepackage{colortbl}
\usepackage{enumitem}
\usepackage{bbm}
\usepackage{array}
\usepackage{microtype}
\definecolor{LightGray}{gray}{0.94}
\definecolor{RankBlue}{HTML}{E8F1FF}
\definecolor{DropGreen}{HTML}{E8F6ED}
\usepackage{xcolor}
\definecolor{thmcolor}{HTML}{0201F5}

\newtheoremstyle{coloredthm}
  {3pt}
  {3pt}
  {\itshape}
  {}
  {\bfseries\color{thmcolor}}
  {.}
  {.5em}
  {}

\theoremstyle{coloredthm}

\newcommand{\vect}[1]{\textbf{#1}}
\newcommand{\mat}[1]{\textbf{#1}}
\newcommand{\set}[1]{\mathcal{#1}}

\definecolor{cvprblue}{rgb}{0.21,0.49,0.74}
\definecolor{supplcolor}{HTML}{A32D26}

\definecolor{noise20color}{HTML}{FFF9E6} 
\definecolor{noise50color}{HTML}{EDF7ED} 
\definecolor{noise80color}{HTML}{EBF5FB} 

\usepackage[breaklinks,colorlinks,citecolor=brown]{hyperref}
\usepackage{thmtools}

\def\paperID{*****} 
\def\confName{CVPR}
\def\confYear{2026}

\newcommand{\method}{EgoAction}
\newcommand{\methodfull}{\textbf{Ego}centric \textbf{A}ction \textbf{c}omposition with reliability-aware \textbf{t}emporal fus\textbf{ion}}

\newcommand{\tious}{\{0.1,0.2,0.3,0.4,0.5\}}

\title{EgoAction: Egocentric Action Composition with Reliability-Aware Temporal Fusion for the EPIC-KITCHENS Action Detection Challenge at CVPR 2026}

\author{Zhiheng Fu$^1$~~Zixu Li$^{1}$~~Zhiwei Chen$^1$~~Fangxu Liu$^{1}$~~Yupeng Hu$^{1}$~~Weili Guan$^{2}$~~Liqiang Nie$^2$ \vspace{2mm}\\
$^1$Shandong University\hspace{1.5cm}$^2$Harbin Institute of Technology (Shenzhen)\hspace{1.5cm}\\
{\tt\small \{fuzhiheng8,lizixu.cs,zivczw,cunyurelax,honeyguan,nieliqiang\}@gmail.com;} \\ 
{\tt\small huyupeng@sdu.edu.cn}\\
}

\begin{document}

\maketitle
\begin{abstract}
The EPIC-KITCHENS-100 Action Detection challenge evaluates whether a model can localize the start and end of each action in long untrimmed egocentric videos and assign the corresponding verb--noun action label. In this report, we formulate our submission as \method{} (\methodfull), a unified decoupled detection and fusion pipeline. The pipeline uses EPIC-finetuned VideoMAE-L features, trains separate noun and verb temporal detectors with causal temporal modeling, composes action hypotheses from top noun--verb pairs, and introduces a confidence-adaptive boundary fusion rule at post-processing time. The key observation is that verb and noun streams often fail differently: verb scores are sensitive to motion transitions, whereas noun scores are sensitive to hand-object visibility and object clutter. A fixed arithmetic mean of their predicted boundaries can therefore amplify localization errors when one stream degenerates. We replace this hard-coded mean with Dynamic Weighted Fusion (DWF), which normalizes the maximum noun and verb classification confidences into proposal-wise boundary weights and linearly combines the two intervals. This lightweight tensor-only operator shifts boundary authority toward the more reliable stream while preserving the decoupled action scoring mechanism. Together with sliding-window inference, top-$K$ noun--verb action composition, and class-wise Soft-NMS, \method{} provides a compact and reproducible system for egocentric temporal action detection.
\end{abstract}

\section{Introduction}
Recent progress in language-conditioned visual reasoning has shown that fine-grained visual understanding often depends on compositional reasoning~\cite{qwenvl,ENCODER,videomae,INTENT,actionformer,PAIR,OFFSET,qwen25vl,EgoAdapt}. EPIC-KITCHENS-100 provides a large-scale benchmark for fine-grained, ego-centric video understanding, containing unscripted kitchen activities recorded by head-mounted cameras across 45 kitchens, 100 hours of video, about 20M frames, 90K action segments, and a vocabulary of 97 verb classes and 300 noun classes~\cite{damen2022rescaling,damen2018scaling,TempRet}. The Action Detection track asks participants to detect every action instance in an untrimmed video and assign a compositional label $(v,n)$, where $v$ is a verb and $n$ is a noun. The official metric is mean Average Precision (mAP) at temporal IoU thresholds from 0.1 to 0.5, reported for verb, noun, and action labels. Consequently, advancing research on such a comprehensive egocentric benchmark holds great potential to facilitate downstream applications in related fields~\cite{OmniEgo-R}, including compositional reasoning and retrieval~\cite{FineCIR,Air-Know,gpt4,MEDIAN,mamba,STABLE,HABIT}, video understanding~\cite{ReTrack,videomambasuite,HUD,internvl,egothink,REFINE}, and multimodal learning~\cite{TEMA,HINT,ERASE,MELT,qwen3technicalreport,ConeSep}.

Unlike trimmed action recognition, this track evaluates the full detection pipeline: a model must search through long untrimmed egocentric recordings, find when each interaction starts and ends, and then assign a valid verb--noun pair. The difficulty therefore does not come from recognition alone, but from the coupling between temporal localization and compositional semantics. A temporally accurate proposal with the wrong noun is counted as wrong; a semantically correct verb--noun pair with a shifted boundary is also penalized, especially at higher tIoU thresholds. This creates a setting where factor-level evidence must be preserved, but the final action prediction must still be fused into a single temporally precise detection.

This observation motivates a decoupled but reliability-aware design. Following the strong OpenTAD family of temporal detectors~\cite{liu2024causaltad,opentad,actionformer,mamba,videomambasuite}, we build on feature-based causal temporal modeling, but adapt it to the 2026 challenge under the available EPIC-finetuned VideoMAE-L features~\cite{videomae}. The original CausalTAD~\cite{liu2024causaltad} report shows that temporal causality is crucial for TAD and reports strong EPIC-KITCHENS action detection results with InternVideo2 features. In our setting, however, the model ought to operate with separate VideoMAE-L noun and verb features and then compose them into action detections. This exposes three concrete challenges:

\textbf{C1: Boundary ambiguity under egocentric motion.} Short interaction phases and head-mounted camera shake create temporally shifted proposals. Even if noun and verb classifiers are correct, their regression heads may localize slightly different start/end points.

\textbf{C2: Stream-specific degeneration.} The noun stream can become uncertain when the target object is small, occluded, or confused with kitchen clutter; the verb stream can become uncertain when the motion is subtle or temporally delayed. Directly averaging their coordinates assumes equal reliability and can pull a correct boundary toward a noisy one.

\textbf{C3: Compositional action sparsity.} EPIC action labels are formed from 97 verbs and 300 nouns, but only a subset of combinations is common. The detector must maintain rich top-$K$ candidates while avoiding a combinatorial explosion before Non-Maximum Suppression (NMS), the post-processing step that removes duplicate temporal proposals with highly overlapping intervals.

To address these challenges, we propose \method{} (\methodfull), a unified decoupled detection and fusion pipeline. Instead of forcing nouns and verbs into a single joint detector or applying a fixed post-processing average, \method{} treats the two semantic factors as complementary streams with different localization reliability. The key idea is to preserve noun--verb factorization through VideoMAE-L evidence extraction and causal temporal detection, and to perform action-level coupling only after each stream has produced proposal-level confidence and boundary estimates. This design allows the model to inherit the temporal modeling strength of CausalTAD while explicitly correcting the failure mode that matters for compositional action detection: when one stream is confident and the other is temporally shifted, the final boundary should not give them equal authority.

Our contributions are threefold, summarized as follows:
\begin{itemize}[leftmargin=*,topsep=2pt,itemsep=1pt]
    \item We analyze EPIC-KITCHENS Action Detection as a coupled localization--composition problem and identify three bottlenecks---egocentric boundary ambiguity, stream-specific degeneration, and compositional action sparsity---that explain why direct joint detection or fixed two-stream fusion is insufficient.
    \item We propose \method{} (\methodfull), a decoupled causal detection pipeline using EPIC-finetuned VideoMAE-L noun and verb features, factor-specific temporal detectors, top-$K$ action composition, and confidence-adaptive Dynamic Weighted Fusion for boundary post-processing.
    \item We achieve 25.94 average action mAP, while maintaining balanced factor-level performance of 28.66 verb mAP and 28.61 noun mAP on the official 2026 EPIC-KITCHENS Action Detection leaderboard.
\end{itemize}
\section{Methodology}

\begin{figure*}[t]
    \centering
    \vspace{-16pt}
    \includegraphics[width=\linewidth]{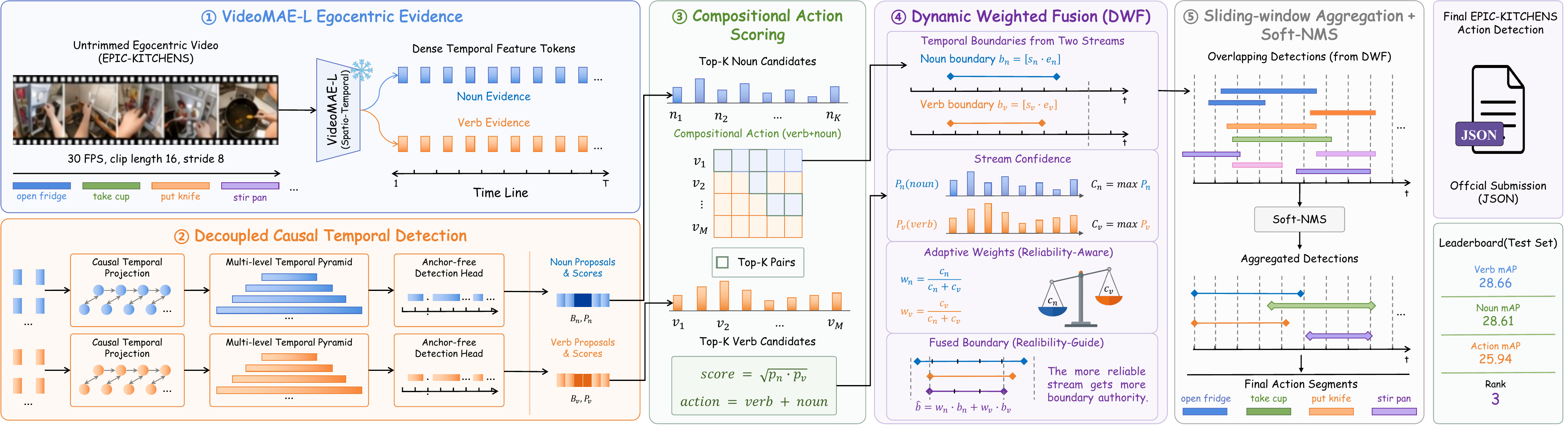}
    \vspace{-22pt}
\caption{Overview of \method. The pipeline preserves noun--verb decoupling until the final compositional stage, where action scores and temporal boundaries are fused with reliability-aware proposal weights.}
        \vspace{-17pt}
    \label{fig:pipeline}
\end{figure*}

As shown in Fig.~\ref{fig:pipeline}, \method{} (\methodfull) is a unified feature-based temporal action detection pipeline built on EPIC-finetuned VideoMAE-L features and CausalTAD-style temporal detectors. The core design is to transform an untrimmed egocentric video into two factor-level proposal streams, one for nouns and one for verbs, and then compose them into action detections only after each stream has produced both semantic confidence and temporal boundaries. This avoids treating the 97$\times$300 action space as a monolithic class set and, more importantly, allows the final temporal boundary to be assigned according to stream reliability rather than a fixed average. In the following, we first formulate the action detection problem and notation, and then elaborate each component of \method{}.

\subsection{Problem Formulation}
Given an untrimmed egocentric video $\set{V}$, temporal action detection predicts a set
\begin{equation}
    \set{Y}=\{(T_i^s,T_i^e,V_i,N_i,S_i)\}_{i=1}^{N},
\end{equation}
where $(T_i^s,T_i^e)$ are scalar start and end times in seconds, $(V_i,N_i)$ are scalar verb and noun labels, and $S_i$ is the scalar detection score. A prediction matches a ground-truth instance if the class is correct and its temporal IoU exceeds a threshold $\tau$:
\begin{equation}
    \mathrm{tIoU}(\vect{b},\hat{\vect{b}})=\frac{|\vect{b}\cap \hat{\vect{b}}|}{|\vect{b}\cup \hat{\vect{b}}|},\qquad \tau\in\tious .
\end{equation}
The challenge score averages AP over classes and thresholds.

\subsection{VideoMAE-L Egocentric Evidence}
We use the pre-extracted VideoMAE-L features provided in OpenTAD~\cite{opentad}. The backbone is VideoMAE-L-16$\times$4$\times$1 finetuned on EPIC-KITCHENS, and features are extracted from 30 FPS videos with clip length 16, snippet stride 8, and frame interval 1. Thus, feature index $i$ corresponds to a local temporal center, formulated as,
\begin{equation}
    \gamma_i=\frac{i\cdot \delta + O}{F},\qquad \delta=8,\quad O=4,\quad F=30,
\end{equation}
where $\delta$ is the snippet stride in frames, $O$ is the scalar offset frame used by the feature window, and $F$ is the scalar frame rate. For long videos, we use sliding-window inference with a maximum window length of 4608 feature steps and 50\% overlap at test time. A predicted feature-domain boundary vector $\vect{b}=(U^s,U^e)$ is converted back to seconds by
\begin{equation}
    \mathrm{sec}(\vect{b})=\frac{\vect{b}\delta + W_0 + O}{F},
\end{equation}
where $W_0$ is the start frame of the current sliding window. This explicit conversion prevents local windows from being interpreted as independent videos.

\subsection{Decoupled Causal Temporal Detection}
The detector is built on the OpenTAD implementation of CausalTAD. For each stream $R\in\set{R}=\{n,v\}$, the feature sequence $\mat{X}^R\in\mathbb{R}^{T\times C}$ is projected into a causal temporal representation and processed by a feature pyramid:
\begin{equation}
    \{\mat{H}_l^R,\mat{M}_l^R\}_{l=1}^{L}=\mathrm{FPN}(\mathrm{CausalProj}(\mat{X}^R,\mat{M}^R)),
\end{equation}
where $\mat{H}_l^R$ and $\mat{M}_l^R$ are the feature and valid-mask matrices at pyramid level $l$. In our configuration, $C=1024$, the projection output dimension is 512, the maximum sequence length is 4608, and $L=7$ pyramid levels use strides $\{1,2,4,8,16,32,64\}$. The anchor-free head predicts per-point class probabilities and distances to segment boundaries:
\begin{equation}
    \mat{P}_l^R=\sigma(g_{\mathrm{cls}}^R(\mat{H}_l^R)),\qquad \mat{D}_l^R=\mathrm{ReLU}(g_{\mathrm{reg}}^R(\mat{H}_l^R)).
\end{equation}
The training objective follows the standard dense TAD loss:
\begin{equation}
    \mathcal{L}^{R}=\lambda_{\mathrm{cls}}\mathcal{L}_{\mathrm{focal}}(\mat{P}^R,\mat{Y}^R)+\lambda_{\mathrm{reg}}\mathcal{L}_{\mathrm{DIoU}}(\mat{D}^R,\mat{B}^R),
\end{equation}
where $\mat{Y}^R$ and $\mat{B}^R$ are assigned class and boundary target matrices. We train the noun and verb detectors separately on the train+val split for the final submission, using AdamW, mixed precision, EMA weights, label smoothing, and cosine learning-rate decay.

\subsection{Cross-Stream Reliability Guidance}
Although the final submission keeps noun and verb outputs decoupled until post-processing, we also implement a dual-stream variant to study feature-level interaction. Its purpose is not to collapse the two tasks, but to expose each stream to the complementary evidence of the other. Given the main-stream feature matrix $\mat{H}_l^R$ and the auxiliary feature matrix $\mat{A}_l^{\bar R}$, we first suppress unreliable auxiliary snippets with an uncertainty gate:
\begin{equation}
    U_t=\sigma(h(\vect{a}_{l,t}^{\bar R})),\quad
    \omega_t=1-\frac{U_t-\min_t U_t}{\max_t U_t-\min_t U_t+\epsilon},
\end{equation}
where $\vect{a}_{l,t}^{\bar R}$ is the auxiliary feature vector at time index $t$, $h(\cdot)$ is a lightweight uncertainty estimator, $\sigma(\cdot)$ is the sigmoid function, $U_t$ is the scalar uncertainty score, $\omega_t$ is the resulting scalar reliability weight, and $\epsilon$ is a small constant for numerical stability. The min--max normalization converts larger uncertainty into smaller reliability. The gated auxiliary feature is then formulated as,
\begin{equation}
    \tilde{\vect{a}}_{l,t}^{\bar R}=\omega_t\vect{a}_{l,t}^{\bar R},
\end{equation}
where $\tilde{\vect{a}}_{l,t}^{\bar R}$ denotes the reliability-weighted auxiliary vector and $\tilde{\mat{A}}_l^{\bar R}$ stacks these vectors over time. A cross-attention update is then applied inside the classification tower:
\begin{equation}
\small
    \mathrm{CWA}(\mat{H}_l^R,\tilde{\mat{A}}_l^{\bar R})=
    \mathrm{MHA}(\mat{Q}=\mat{H}_l^R,\mat{K}=\tilde{\mat{A}}_l^{\bar R},\mat{V}=\tilde{\mat{A}}_l^{\bar R}),
\end{equation}
where $\mathrm{MHA}$ denotes multi-head attention, and $\mat{Q}$, $\mat{K}$, and $\mat{V}$ are the query, key, and value matrices constructed from the main and gated auxiliary streams. The updated main-stream feature is formulated as,
\begin{equation}
    \mat{H}_l^R \leftarrow \mat{H}_l^R+\mathrm{CWA}(\mat{H}_l^R,\tilde{\mat{A}}_l^{\bar R}).
\end{equation}
This cross-stream module clarifies the same principle used by the final model: noun evidence and verb evidence are complementary, but their reliability varies over time. We therefore keep the submitted detector decoupled for robustness and move the most explicit reliability decision to the post-processing stage. The next two subsections describe this transition: first composing factor-level predictions into actions, then assigning boundary authority through DWF.

\subsection{Compositional Action Scoring}
For each aligned proposal index $j$, the noun detector outputs a boundary vector $\vect{b}_j^n$ and class-score vector $\vect{p}_j^n\in\mathbb{R}^{300}$, while the verb detector outputs $\vect{b}_j^v$ and $\vect{p}_j^v\in\mathbb{R}^{97}$. We keep the top $K_n=10$ noun classes and top $K_v=10$ verb classes for each proposal. A composed action hypothesis $(p,q)$ receives the geometric-mean score
\begin{equation}
    S_{j,p,q}=\sqrt{P_{j,p}^{n}P_{j,q}^{v}},\qquad
    A_{p,q}=300q+p,
\end{equation}
where $p$ is a noun index, $q$ is a verb index, and $P_{j,p}^{n}$ and $P_{j,q}^{v}$ denote scalar entries of the noun and verb score vectors. The geometric mean keeps the action score conservative: a candidate is strong only when both object and motion evidence support it. Yet the score alone does not resolve which stream should determine the start and end time. This motivates the boundary-specific fusion rule in the next subsection.

\subsection{Dynamic Weighted Fusion}
The original two-stream post-processing used the hard arithmetic mean
\begin{equation}
    \vect{b}_j^{\mathrm{mean}}=\frac{1}{2}(\vect{b}_j^n+\vect{b}_j^v).
\end{equation}
This is unbiased only when both streams have comparable localization reliability. In practice, one stream can become unreliable because of occluded nouns, ambiguous motion, or a shifted interaction phase. The mean then moves the correct boundary toward the degraded stream.

Thus, we design DWF to estimate proposal-level reliability from classification confidence. Specifically, for each proposal $j$, we compute scalar stream confidences, formulated as,
\begin{equation}
    C_j^n=\max_{p}P_{j,p}^{n},\qquad C_j^v=\max_{q}P_{j,q}^{v}.
\end{equation}
The normalized stream weights are
\begin{equation}
    W_j^n=\frac{C_j^n}{C_j^n+C_j^v+\epsilon},\qquad
    W_j^v=\frac{C_j^v}{C_j^n+C_j^v+\epsilon},
\end{equation}
where $\epsilon=10^{-6}$ prevents division by zero. The fused boundary is
\begin{equation}
    \hat{\vect{b}}_j=W_j^n\vect{b}_j^n+W_j^v\vect{b}_j^v.
    \label{eq:dwf}
\end{equation}
Equation~\eqref{eq:dwf} is applied to both start and end coordinates. It changes only the temporal interval associated with proposal $j$; the semantic action score remains $S_{j,p,q}=\sqrt{P_{j,p}^{n}P_{j,q}^{v}}$. Thus, DWF does not bias the action classifier toward nouns or verbs. Instead, it assigns boundary authority to the stream that appears more reliable for the current proposal, with only two reductions, a normalization, and a vectorized linear interpolation. After DWF, the composed action labels, geometric-mean scores, and fused intervals are filtered by confidence, truncated to the top-scoring candidates, and finally passed to class-wise Soft-NMS across sliding-window outputs to suppress duplicate detections before JSON conversion.

\section{Experiments}
\subsection{Challenge Setup}
The 2026 EPIC-KITCHENS Action Detection track is evaluated on the hidden test split through Codabench. Each submission returns a JSON dictionary of detections for untrimmed videos. The leaderboard reports verb, noun, and action mAP at $\tau\in\tious$ and their averages. The final \method{} submission was with supervision-level fields PT=2, TL=3, and TD=4 as recorded by the leaderboard.

\subsection{Implementation Details}
Following~\cite{liu2024causaltad}, the noun and verb models are trained separately with train+val annotations for the final test submission. We use batch size 16 for the single-stream CausalTAD configuration, AdamW with learning rate $10^{-4}$ and weight decay 0.05, gradient clipping at 1, EMA, AMP, 5 warm-up epochs, and 50 total epochs for the noun/verb detector configuration. The post-processing stage keeps up to 5000 candidates before NMS and at most 3000 detections per video. Noun post-processing uses Soft-NMS with $\sigma=0.6$, minimum score 0.005, and voting threshold 0.65, while verb/action post-processing uses $\sigma=0.4$, minimum score 0.001, and voting threshold 0.75.

\subsection{Official Leaderboard Results}
Table~\ref{tab:leaderboard} reports the official public leaderboard around our entry. Our method ranks third in the Action Detection track with 25.94 action average mAP among all submission. It is very close to the second-ranked system, trailing by 0.31 action-mAP points, while outperforming the fourth-ranked entry by 1.73 points. The result is obtained with VideoMAE-L features and the DWF post-processing pipeline, without an explicit multi-seed ensemble.

\begin{table*}[t]
\centering
\small
    \vspace{-16pt}
\caption{Official Codabench results for the EPIC-KITCHENS-100 Action Detection track. Scores are mAP (\%). ``Avg.'' averages thresholds 0.1 to 0.5. The highlighted row is the final \method{} submission.}
\vspace{-8pt}
\label{tab:leaderboard}
\resizebox{0.82\linewidth}{!}{
\begin{tabular}{cllccccccccc}
\toprule
Rank & Method & Type & \makecell{Verb\\Avg.} & \makecell{Noun\\Avg.} & \makecell{Action\\0.1} & \makecell{Action\\0.2} & \makecell{Action\\0.3} & \makecell{Action\\0.4} & \makecell{Action\\0.5} & \makecell{Action\\Avg.}\\
\midrule
1 & KAUST-4Paradigm-MoonshotAI-Nvidia & Official Baseline & 30.02 & 35.22 & 36.10 & 34.70 & 32.67 & 29.91 & 26.50 & 31.98\\
2 & dg\_team(deepglint) & Official Baseline & 26.87 & 29.56 & 29.29 & 28.33 & 26.80 & 24.77 & 22.06 & 26.25\\
\rowcolor{RankBlue}
3 & \textbf{\method{}} & 2026 New Submission & \textbf{28.66} & \textbf{28.61} & \textbf{29.56} & \textbf{28.45} & \textbf{26.50} & \textbf{24.34} & \textbf{20.84} & \textbf{25.94}\\
4 & Oxford+Bristol & Official Baseline & 27.12 & 29.36 & 28.13 & 26.74 & 25.01 & 22.29 & 18.86 & 24.21\\
5 & yy & 2026 New Submission & 24.13 & 23.70 & 22.70 & 21.85 & 20.53 & 18.58 & 16.22 & 19.98\\
6 & CY & 2026 New Submission & 3.40 & 3.00 & 3.63 & 3.07 & 2.41 & 1.65 & 0.97 & 2.35\\
\bottomrule
\end{tabular}
}
\vspace{-17pt}
\end{table*}

\subsection{Analysis of Noun--Verb Balance}
The final result shows balanced verb and noun detection averages: 28.66 and 28.61 mAP, respectively. However, the action score is 25.94, about 2.7 points lower than each factor alone. This gap is expected because an action match requires correct temporal localization and both semantic factors. It also justifies maintaining decoupled noun and verb streams until the last stage: a joint detector would need to learn a much larger and sparser action vocabulary, while our composition rule reuses the stronger factor-level predictions.

The threshold breakdown reveals that localization precision remains the main bottleneck. Our action mAP decreases from 29.56 at tIoU 0.1 to 20.84 at tIoU 0.5. DWF directly targets this regime: if a noun proposal is confident but a verb proposal is shifted by a delayed motion cue, Eq.~\eqref{eq:dwf} assigns more authority to the noun boundary; conversely, when an object is partially visible but the motion phase is clear, the verb boundary dominates. Thus, DWF is valuable for medium and high tIoU thresholds, where a small boundary shift can change a true positive into a false positive.

\subsection{Diagnostic Component Study}
Hidden test labels prevent a strict one-variable ablation of the final submission. We therefore report a diagnostic component study in Table~\ref{tab:diagnostic}, separating quantities that are directly evaluated from mechanism-level changes. The key change from the original post-processing is replacing the hard mean with DWF. This change is deterministic, class-agnostic, and doesn't alter trained detectors or classification scores.

\begin{table}[t]
\centering
\scriptsize
\setlength{\tabcolsep}{3.5pt}
\caption{Diagnostic component study. Numeric rows are evaluated records; mechanism rows describe the exact post-processing change when hidden labels do not permit isolated scoring.}
\vspace{-8pt}
\label{tab:diagnostic}
\resizebox{0.86\linewidth}{!}{
\begin{tabular}{p{0.35\linewidth}p{0.25\linewidth}p{0.30\linewidth}}
\toprule
Component & Evidence & Effect\\
\midrule
Separate noun detector & val noun mAP 36.25 & strong object stream\\
Separate verb detector & val verb mAP 32.91 & strong motion stream\\
Hard joint composition & val action mAP 28.86 & exposes fusion bottleneck\\
\rowcolor{DropGreen}
DWF boundary fusion & Eq.~\eqref{eq:dwf} & adaptive boundary authority\\
Top-$10\times10$ composition & implemented & controls action sparsity\\
Soft-NMS voting & implemented & suppresses duplicate windows\\
\bottomrule
\end{tabular}
}
\vspace{-12pt}
\end{table}

\subsection{Why Confidence is a Useful Boundary Prior}
DWF is based on a simple but useful assumption: for a dense anchor-free detector, the stream that assigns a sharper semantic posterior to a proposal is often also the stream whose local temporal evidence is cleaner. This is not a learned calibration theorem, but it follows the training coupling between classification and localization. For a proposal $j$, let $E_j^R$ denote the unknown scalar boundary error of stream $R$. If the confidence-derived weight is positively correlated with inverse error, then DWF reduces the expected absolute boundary error compared with an unconditioned mean:
\begin{equation}
    \mathbb{E}\big[|\hat{\vect{b}}_j-\vect{b}_j^\star|\big]
    \leq \mathbb{E}\big[|\tfrac{1}{2}(\vect{b}_j^n+\vect{b}_j^v)-\vect{b}_j^\star|\big]
\end{equation}
whenever the more confident stream has smaller expected error by a margin larger than the weight-estimation noise. Here $\vect{b}_j^\star$ is the ground-truth boundary vector. This explains why the operator is particularly relevant for EPIC-KITCHENS: noun confidence drops under occlusion and clutter, while verb confidence drops under subtle or delayed motion. A fixed average cannot express this sample-level asymmetry.

DWF also preserves the semantics of the original two-stream detector. The class score remains the geometric mean $S=\sqrt{P^nP^v}$, so an action still needs both object and motion support. Only the temporal coordinate is reassigned according to reliability. This separation is important: increasing the score of the high-confidence stream alone would over-favor either nouns or verbs, while Eq.~\eqref{eq:dwf} changes only the localization decision.

\subsection{Failure Modes and Future Directions}
The main remaining failure mode is high-tIoU localization under very short interactions. In such cases both streams may be confident but shifted in the same direction because the VideoMAE-L snippet stride discretizes the transition. Another failure mode is semantic co-occurrence bias: frequent actions such as open/fridge or take/cup can survive top-$K$ composition even when the visual noun is ambiguous. Finally, the current DWF uses only maximum class confidence, which is intentionally cheap but does not explicitly measure posterior entropy or stream disagreement.

Future improvements could therefore replace the scalar confidence with a richer reliability vector, e.g., entropy, temporal sharpness, and noun--verb consistency. A learned calibration head could predict $W_j^n,W_j^v$ from the two posterior score vectors and the regressed interval disagreement $|\vect{b}_j^n-\vect{b}_j^v|$. We did not introduce such learned calibration in the final submission to avoid overfitting to validation labels and to keep the challenge pipeline deterministic and robust.

\section{Conclusion}
We presented \method{} (\methodfull), which combined EPIC-finetuned VideoMAE-L features, decoupled causal temporal detectors, top-$K$ noun--verb action composition, sliding-window aggregation, and a new Dynamic Weighted Fusion rule for boundary post-processing. The central lesson is that two-stream action detection should not only combine scores; it should also decide which stream is more trustworthy for temporal localization. DWF provides this decision with negligible overhead by converting classification confidence into proposal-wise boundary weights. Our final submission achieves 25.94 average action mAP on the public leaderboard.

{
    \small
    \bibliographystyle{ieeenat_fullname}
    \bibliography{main}
}

\end{document}